\documentclass[11pt]{article}
\usepackage[blue]{jhu_paper}

\date{}
\usepackage{array}
\usepackage{colortbl}
\usepackage{tabularx}
\usepackage{multirow}
\usepackage{subcaption}
\usepackage{enumitem}
\usepackage{wrapfig}
\usepackage[numbers,sort&compress]{natbib}
\usepackage{needspace}

\definecolor{BestCellBlue}{RGB}{232,244,255}
\definecolor{SecondCellBlue}{RGB}{244,249,255}

\title{Track, Articulate, Act}
\jhusubtitle{Generating Articulation from Casual Human Videos}
\author{Jiaming Zhang \qquad Homanga Bharadhwaj\\[0.2em]
}
\jhuinstitution{Department of Computer Science, Johns Hopkins University}
\jhucontact{\texttt{jzhan282@jhu.edu} \quad \texttt{homanga@jhu.edu}}
\jhuvenue{Brains, Bots, and Behavior Lab}
\jhurunningtitle{Track, Articulate, Act}

\makeatletter
\providecommand{\@LN}[2]{}
\providecommand{\@LN@col}[1]{}
\makeatother
\begin{document}
\maketitle

\begin{figure*}[h!]
    \centering

    \makebox[\textwidth][c]{%
        \small
        \href{https://track-articulate-act.github.io/}
        {\texttt{\textbf{track-articulate-act.github.io}}}
    }
    \vspace{0.3em}

    \includegraphics[width=\linewidth]{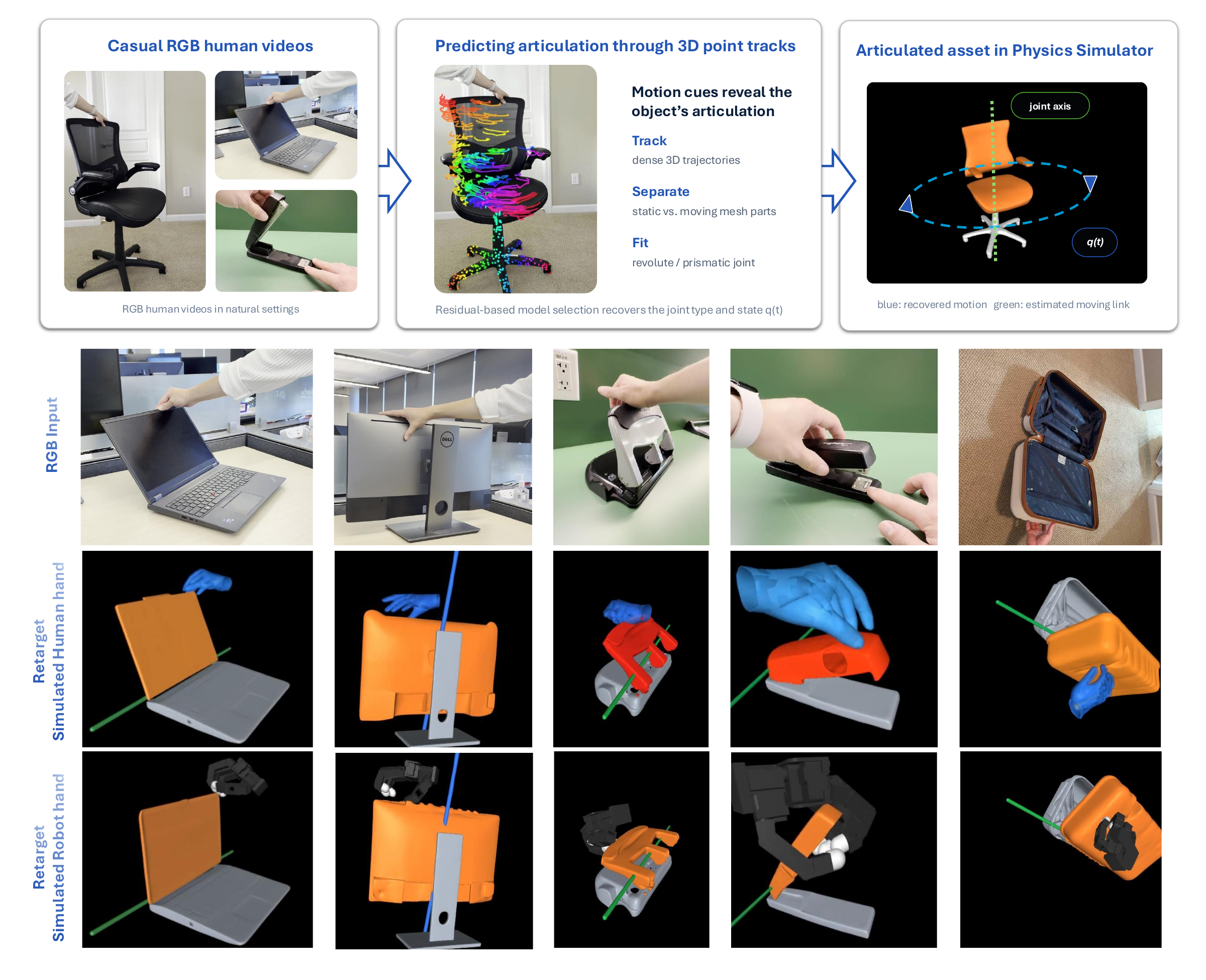}
    \caption{\footnotesize \textbf{Overview}.
        Starting from only a casual RGB human video,
        our method extracts motion cues, predicts articulation, constructs an approximate simulation-ready articulated object model, and replays the recovered hand-object interaction in a physics simulator. Results show that our framework is able to infer articulation for different types of joints including revolute and prismatic across diverse real-world objects.}
    \label{fig:teaser}
\end{figure*}

\begin{abstract}
Human videos contain rich causal evidence for robot manipulation: they reveal how hand motion induces object motion and produces task-relevant changes in object state. In this work, we study articulated objects such as doors, drawers, cabinets, laptops, ovens, and hinged containers that are ubiquitous in daily life and present unique challenges for embodied interaction. These objects cannot be represented by a single pose; their motion depends on the underlying parts and joints. We introduce a real-to-sim framework that reconstructs a simulation-ready articulated object and hand--object interaction from a casual monocular RGB video, without RGB-D or multi-view input, prior scans, manually specified joints, or robot demonstrations. Our key insight is that dense 3D point tracks provide an embodiment-agnostic articulation cue: points on the fixed link remain approximately stationary, while points on the moving link follow coherent revolute or prismatic motion. Our method segments the links, estimates the joint and its state trajectory, reconstructs an articulated asset, and aligns the recovered 3D hand motion with the object. Central to our approach is a modular recipe that repurposes powerful pretrained models for single-image 3D reconstruction, mesh segmentation, and 3D scene flow, connecting their predictions through explicit geometric reasoning to infer articulation. We use the reconstructed articulated object and the human hand trajectory to replay interactions through contact in MuJoCo. The framework shows how pretrained vision models and explicit motion reasoning can turn casual human videos into articulated object models suitable for downstream embodied interactions.
\end{abstract}


\section{Introduction}
\label{sec:intro}

Everyday environments are filled with objects that open, slide, rotate, and fold. Cabinets, drawers, laptops, box lids, ovens, and hinged containers are central to household, office, and workshop activities. Interacting with them is a fundamental challenge for embodied intelligence: successful action requires understanding which part moves, which part remains fixed, how they are connected, and how contact changes the object state. Human interactions reveal this underlying structure very naturally. We study how to recover these embodiment-agnostic motion cues from ordinary human videos and use them for articulated object manipulation.

We ask whether this structure can be recovered from a single casually captured monocular video of a person manipulating an articulated object. Although easy to collect at scale, such a video provides no actions, depth observations, part labels, joint annotations, object meshes, or simulator-ready assets. Recent work has shown the value of converting human videos into explicit 3D, object-centric, and interaction-centric representations such as point tracks, hand--object trajectories, and scene flow~\cite{bharadhwaj2024track2act,guzey2025aina,chen2026videomanip,li2025novaflow,bharadhwaj2026motionforesight}. We build on this principle, but must recover both the observed motion and the kinematic structure that explains it. A drawer front translates relative to a static frame, whereas a laptop screen or cabinet door rotates about a hinge. Unlike rigid-object pose, articulated-object pose is therefore not well-defined until the parts, joints, and moving link are identified~\cite{wen2024foundationpose,lee2025any6d}.

When a person opens a lid or pulls a drawer, points on the moving link follow a coherent low-dimensional trajectory, while points on the static structure remain approximately fixed. Dense 3D tracks therefore provide an interface between general video perception and explicit kinematic reasoning: their spatial grouping suggests the part decomposition, and their trajectories distinguish translation along an axis from rotation about a hinge. Recent progress in long-range tracking and monocular 4D reconstruction makes these cues increasingly accessible in ordinary videos~\cite{karaev2024cotracker,karaev2025cotracker3,xiao2025spatialtrackerv2,karhade2025any4d,harley2025alltracker}. The remaining challenge is to turn noisy tracks and incomplete visual observations into coherent link geometry, a joint model, and a metrically aligned interaction.

Recent work has made substantial progress on articulation recovery from videos, using learned part-and-joint predictors, dynamic reconstruction, point-track optimization, or geometric primitives~\cite{artykov2025sim2art,liu2025videoartgs,guo2026articulat3d,artykov2026articulationprime}. Complementary systems reconstruct articulated hand--object interactions by coordinating geometry, pose, and contact priors~\cite{wang2026arthoi,liu2026claytostone}. These methods establish that casual videos can support rich articulated reconstruction. Their primary focus, however, is recovering geometry and kinematics, typically evaluated through joint, reconstruction, tracking, or rendering accuracy. We study a complementary research question: whether the recovered scene can be instantiated as an explicit collision asset in physics simulators and whether the observed human hand motion can physically produce the demonstrated state change. \textit{This requires not only plausible geometry and visual articulation, but also consistent scale, hand--object alignment, joint limits, and contact behavior.}

This goal naturally suggests a \textit{modular approach}: although no single pretrained model recovers a simulation-ready articulated interaction, current models offer complementary estimates of object geometry, motion, and hand--object interaction. Starting from a casual monocular video, we select low-occlusion frames that span the observed articulation and reconstruct a mesh hypothesis from each. We use part masks to transfer the static--moving decomposition onto these meshes and register them through the static link, bringing the geometry into a common object frame. Within this frame, we ground dense 3D tracks using the same monocular geometry and fit multiple revolute and prismatic joint hypotheses to the observed moving-link trajectories. We refine and select the joint parameters and type by requiring the rendered link motion to agree with both the tracked points and the observed silhouettes. Undoing the recovered articulation then allows us to fuse the multi-frame link observations into a canonical articulated asset. Finally, we align the reconstructed 3D hand trajectory with the object and replay the interaction in a physics simulator (MuJoCo). 

In this way, \textit{we connect predictions from general-purpose pretrained models through geometric reasoning and optimization}, without training an articulation-specific model. This modular design \textit{allows the framework to benefit from advances in its individual components} while preserving clear geometric and physical assumptions. In summary, our contributions are:

\begin{enumerate}[leftmargin=*]
    \item We develop a modular real-to-sim framework that converts a casual monocular human video into a simulation-ready articulated object and aligned hand--object interaction. The framework connects pretrained models for segmentation, single-image 3D reconstruction, monocular geometry, dense tracking, and hand reconstruction without training an articulation-specific model.
    \item We propose a motion-grounded optimization for articulation inference that registers multi-frame object reconstructions, fits revolute and prismatic joint hypotheses to mesh-grounded 3D point trajectories, and jointly refines the joint parameters and state trajectory through track and silhouette re-projection objectives.
    \item We enable interaction with the reconstructed object in physics simulation by constructing explicit link and collision geometry, aligning the recovered 3D hand trajectory with the articulated asset, and re-targeting the interaction to a simulated hand whose contact drives the passive object joint.
\end{enumerate}


\section{Related Work}
\label{sec:related}

\paragraph{Learning manipulation with human videos}
Human videos offer manipulation experience at a scale that robot data cannot yet match, but they do not directly provide actions, metric 3D object state, or contact supervision. Prior work has used these videos to learn representations and rewards~\cite{nair2022r3m,ma2022vip}, imitate human play or demonstrations~\cite{wang2023mimicplay,shi2025zeromimic,jain2024vid2robot,tang2025mimicfunc}, and extract transferable cues such as affordances, hand poses, point trajectories, and future object motion~\cite{agarwal2023dexterousfunctionalgrasping,bahl2023affordances,bharadhwaj2024track2act,li2025novaflow,soraki2026objectforesight}. Recent approaches also use generated videos and learned world models to provide visual plans and action priors~\cite{liang2025videopolicy,patel2025imitatinggeneratedvideos,chen2025largevideoplanner,goswami2025dexwm,huang2026pointworld, ranawaka2026simfoundry}. Richer capture systems, including smart glasses, headsets, motion capture, and calibrated robot scenes, make human motion easier to transfer but introduce additional sensing and alignment assumptions~\cite{guzey2025aina,wang2024dexcap,chen2026videomanip}. We instead consider the setting of an ordinary monocular video to recover the articulated object model required for interaction: its static and moving links, joint type and parameters, state trajectory, and alignment with the human hand.

\paragraph{Hand--object reconstruction from video}
Recent progress in monocular 3D reconstruction makes it possible to recover complementary elements of a hand--object interaction from ordinary video. Hand models estimate pose and shape from individual images~\cite{rong2020frankmocap,pavlakos2024hamer,potamias2024wilor}, while video-based methods recover temporally consistent hand motion in a global frame~\cite{ye2025haptic,zhang2025hawor}. In parallel, single-image object reconstruction models such as SAM 3D Objects can infer complete mesh hypotheses from partial visual observations~\cite{sam3d}. Joint hand--object approaches further reason about object geometry, pose, and contact, although most focus on rigid objects~\cite{hasson2019jointreconstruction,cao2021reconstructinghoi,ye2023diffhoi,fan2024hold,liu2024easyhoi}. ArtHOI and Clay-to-Stone extend this setting to articulated interactions by coordinating geometric, motion, and contact priors~\cite{wang2026arthoi,liu2026claytostone}. Our framework builds directly on these advances: we use SAM 3D Objects to reconstruct object geometry and HaWoR to recover the global 3D hand trajectory. We then connect these predictions with dense motion tracks and explicit kinematic reasoning to recover a low-DoF simulator joint, align the hand with the resulting collision geometry, and test whether replaying its trajectory physically actuates the passive joint.

\paragraph{Articulated object reconstruction}
Articulation recovery has traditionally relied on explicit 3D observations or multiple views. Classical kinematic fitting and later reconstruction systems recover rearticulable models from rigid-part motion, interaction point clouds, dynamic 4D point clouds, two-state multiview images, or multiview videos~\cite{sturm2011kinematic,jiang2022ditto,liu2023reart,liu2023paris,noguchi2022watchitmove}. RSRD similarly begins from a static multiview scan~\cite{kerr2024rsrd}, while Video2Articulation and ArtiPoint use RGB-D video, with ArtiPoint explicitly handling camera motion and partial visibility through point tracking and factor-graph optimization~\cite{peng2025video2articulation,werby2025artipoint}. These settings provide stronger geometric observations than the casual monocular videos we consider in this work.

Recent work relaxes these assumptions in several complementary ways. Sim2Art learns monocular part and joint prediction from synthetic training data~\cite{artykov2025sim2art}; VideoArtGS and Articulat3D combine video motion cues with video-specific 3D reconstruction and explicit joint fitting~\cite{liu2025videoartgs,guo2026articulat3d}; and Articulation in Prime fits visibility-aware geometric primitives without relying on long-term correspondences~\cite{artykov2026articulationprime}. REACTO instead targets general rearticulable surface reconstruction without restricting the output to an explicit revolute or prismatic simulator joint~\cite{song2024reacto}. Another line of work generates articulated assets from language, images, or independently reconstructed parts, rather than recovering the articulation demonstrated in a video~\cite{mandi2024real2code,le2024articulateanything,qiu2025articulateanymesh,li2025art}.

Compared with existing methods that recover articulated objects from casual monocular videos~\cite{liu2025videoartgs,guo2026articulat3d}, we target a \textit{simulation-ready hand--object interaction} rather than object reconstruction alone. Instead of training an articulation-specific model, we connect predictions from replaceable, general-purpose vision models through multi-frame registration, dense tracking, and explicit low-dimensional kinematic fitting. This modular design allows inheriting improvements in the respective pretrained components that have wide community interest. We additionally reconstruct the 3D hand trajectory and align it with the recovered object and its collision geometry. The resulting links and passive joint are instantiated in MuJoCo, enabling interactive physics simulation and allowing us to test whether contact with the replayed hand produces the demonstrated state change. This physical evaluation complements the emphasis of prior work on joint, geometry, tracking, and rendering accuracy.

\begin{figure*}[t]
    \centering
    \includegraphics[width=\linewidth]{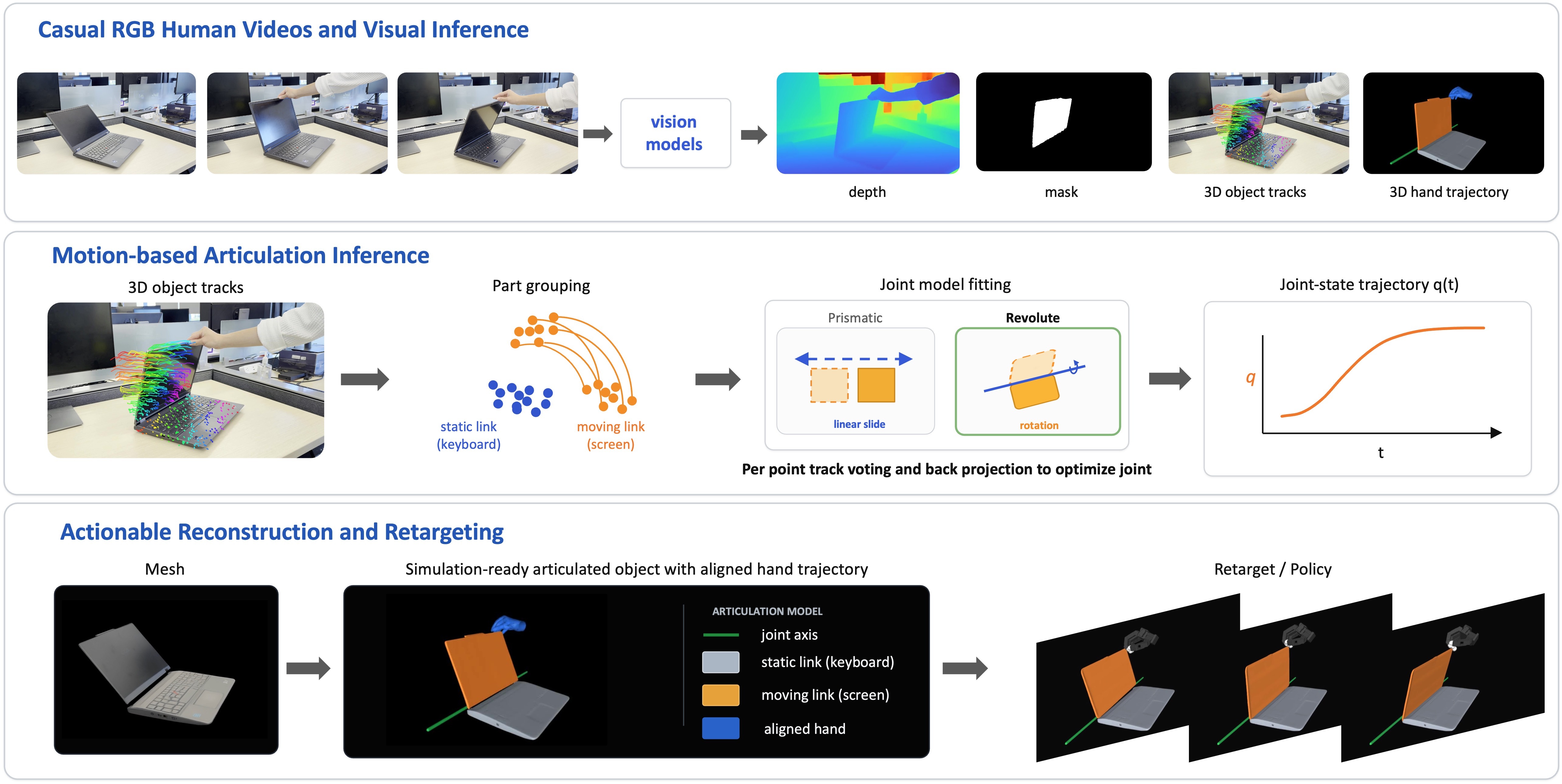}
    \caption{\textbf{Method.} Given a casual human manipulation video, we select frames that expose different articulation states, reconstruct and register part-level geometry, and fit joint hypotheses to metrically grounded 3D tracks. We refine the joint through track and silhouette reprojection, then align and replay the recovered 3D hand motion with the articulated object in MuJoCo.}
    \label{fig:method}
\end{figure*}

\section{Problem Formulation}
\label{sec:problem}

We consider a monocular RGB video $V=\{I_t\}_{t=1}^{T}$ of a human manipulating an articulated object. From this video alone, we recover
\begin{equation}
    \mathcal{O}=(\mathcal{M}_s,\mathcal{M}_m,\mathcal{J},q_{1:T}),
\end{equation}
where $\mathcal{M}_s$ and $\mathcal{M}_m$ are the static and moving links, $\mathcal{J}$ is a revolute or prismatic joint, and $q_{1:T}$ is the demonstrated joint trajectory. We also recover the 3D human hand motion and align it with the object in simulation. We assume one dominant joint per object, but no sensor depth, object scan, CAD model, part label, or joint annotation. Unlike rigid-object pose estimation, the object pose is not defined until its links and their kinematic constraint are known; our goal is therefore to infer the model that explains the observed motion.


\section{Track, Articulate, Act}
\label{sec:method}

Our method turns a monocular human video into a simulation-ready articulated hand--object scene without training an articulation-specific model. We use pretrained models to recover temporally consistent masks, single-frame mesh hypotheses, monocular geometry, dense point correspondences, and global hand motion. We then connect these predictions through four explicit geometric interfaces. First, we select low-occlusion frames, reconstruct each independently, and register the resulting meshes through the static link (Sec.~\ref{sec:multiframe_geometry}). Second, we place dense tracks in the same object frame using the shared monocular geometry and reconstructed surfaces (Sec.~\ref{sec:tracks}). Third, we fit revolute and prismatic joint hypotheses and refine them by requiring the rendered motion to agree with the observed tracks and silhouettes (Secs.~\ref{sec:joint_hypotheses}--\ref{sec:joint_refinement}). Finally, we align the recovered 3D hand motion with the articulated object and re-target it open-loop in MuJoCo to a simulated human hand model, where contact with the hand drives the passive object joint (Sec.~\ref{sec:retargeting}).

\subsection{Object Masks and Multi-Frame Reconstruction from Video}
\label{sec:multiframe_geometry}

Hands in videos often occlude precisely the link that moves, making an arbitrary video frame a poor source of object geometry. We use SAM 3~\cite{carion2025sam3} to track the object and hand masks, choose the least-occluded valid frame as the reference, and add a few low-occlusion keyframes that span large object-track displacement. These frames typically capture open, intermediate, and closed configurations. We project the static and moving point clusters described in Sec.~\ref{sec:tracks} into a clear frame and use them as point prompts to obtain the corresponding part masks. We then propagate these masks through the video, remove hand pixels, and assign every object pixel to at most one link.

At each keyframe, we estimate depth and camera geometry with Depth Anything 3 (DA3)~\cite{lin2025depthanything3} and reconstruct an object mesh from the crop and mask using SAM 3D~\cite{sam3d}. We do not assume access to ground-truth sensor depth. We transfer the propagated static--moving decomposition to each mesh by guiding SegviGen~\cite{li2026segvigen} with the corresponding part masks. Projected vertex support across keyframes resolves ambiguous labels and keeps the two link meshes disjoint.

Because these meshes are reconstructed independently, they vary in shape, scale, and pose. We ground each prediction with the estimated depth and camera intrinsics, then register the keyframes through the link known to be static. The alignment uses a robust depth-aware similarity transform that takes into account any residual scale inconsistency and variability in single-image object reconstructions. We fuse the registered static observations into $\mathcal{M}_s$ and retain the moving-link meshes in their observed configurations. Their relative displacement provides evidence for the joint; once the joint is recovered, we undo this motion and fuse them into the canonical moving link $\mathcal{M}_m$.

\subsection{Metric 3D Point Tracks}
\label{sec:tracks}

We recover dense reference-anchored correspondences and visibility over the complete video with TrackCraft3R~\cite{nam2026trackcraft3r}. Tracking operates directly on the full RGB frames and does not require object, part, or hand masks. After tracking, we use the propagated whole-object and hand masks only to retain object trajectories and discard hand and background trajectories. Let $\mathbf{u}_{i,t}$ be the image location of track $i$ at time $t$. Using depth $D_t$, intrinsics $\mathbf{K}_t$, and a transform $\mathbf{G}_t$ from the camera to the canonical object frame, we first lift it to
\begin{equation}
    \widetilde{\mathbf{p}}_{i,t}
    =\mathbf{G}_t\!\left[
    D_t(\mathbf{u}_{i,t})\mathbf{K}_t^{-1}
    \overline{\mathbf{u}}_{i,t}\right],
\end{equation}
where $\overline{\mathbf{u}}_{i,t}$ is the homogeneous image coordinate. DA3 provides the initial depth and camera parameters. We separate the resulting camera-compensated trajectories into static and moving point clusters according to their motion: static points remain approximately fixed, whereas moving points follow a coherent motion. Projecting these clusters into the video provides the point prompts used to obtain the part masks in Sec.~\ref{sec:multiframe_geometry}.

The tracker provides useful correspondences, but its framewise 3D coordinates can inherit some monocular depth fluctuations. After registering the reconstructed meshes through the static link, we refine $\mathbf{G}_t$ by aligning the observed static surface across frames. We then retain the tracker’s image correspondences and, at the selected keyframes, replace the lifted depth with the first intersection between the corresponding camera ray and the registered mesh surface. We denote the resulting mesh-grounded points by $\mathbf{p}_{i,t}$. Using the same monocular geometry for mesh registration and track grounding keeps their coordinate frames and scales compatible. 

\subsection{Joint Hypotheses from Point Trajectories}
\label{sec:joint_hypotheses}

We explain the moving-link trajectories with a single low-dimensional kinematic model. A prismatic candidate shares one translation direction across time, whereas a revolute candidate shares one axis and pivot. For joint class $y$, parameters $\boldsymbol{\theta}_y$, state trajectory $q_{1:T}$, and visibility $v_{i,t}$, we minimize
\begin{equation}
    E_{\mathrm{track}}(y)=
    \sum_{i,t}v_{i,t}\,
    \rho\!\left(\left\|\mathbf{p}_{i,t}-
    g_y(q_t;\boldsymbol{\theta}_y)\mathbf{p}_{i,r}
    \right\|_2^2\right)
\end{equation}
over moving-link tracks, where $g_y$ is the corresponding translation or rotation and $\rho$ is a robust loss. We initialize prismatic candidates from the dominant displacement and revolute candidates from the observed velocity field.  We built multiple subsets by sampling from the mesh-grounded points at key frames using RANSAC to produce several hypotheses, which is important when the motion is small or some depths are incorrect. The fit also recovers the scalar joint trajectory $q_{1:T}$, which we smooth over time while preserving its endpoints. $E_{\mathrm{track}}$ is calculated per joint type, and the actual joint type is determined by $ y^*=\min \{E_{\mathrm{track}} (y); \mid y \in \{ \text{prismatic}, \text{revolute}\}\} $

\subsection{Joint Refinement by Reprojection}
\label{sec:joint_refinement}

A candidate can fit noisy 3D tracks while still moving the reconstructed link outside the observed object. We therefore replay every candidate joint hypothesis with the reconstructed mesh, which is articulated about the candidate joint with an estimated joint state $q_{1:T}$, and back-project the mesh onto the image plane. We jointly refine the joint parameters and $q_{1:T}$ using intersection over union (IoU) between the back-projected mesh and segmentation mask. A weak temporal penalty discourages abrupt changes in joint state. We select the candidate with the smallest normalized objective over the full sequence, requiring the final joint to explain both the point motion and the visible object shape.

There could be cases when a revolute joint located infinitely distant from the mesh could produce the same motion as a prismatic joint, which will result in a nearly identical cost between two joint types, i.e. $\|E_{\mathrm{track}}(\text{revolute}) - E_{\mathrm{track}}(\text{prismatic})\| \leq \tau $.

As a tie-breaker, we introduce a PCA-based naive joint estimation that extracts the principal axes of the moving link at key frames and consider it as a prismatic joint only when the PCA axes have less than 5$^\circ$ of variance across the articulation.
Once the joint is fixed, we transform every moving-link reconstruction back to a common link frame and fuse the aligned observations into $\mathcal{M}_m$. The range of the recovered trajectory defines the joint limits, and we generate simple collision geometry for both links. The resulting MuJoCo asset contains the static and moving meshes, a single revolute or prismatic joint, and the demonstrated state trajectory.

\begin{figure*}[t]
    \centering
    \includegraphics[width=\linewidth]{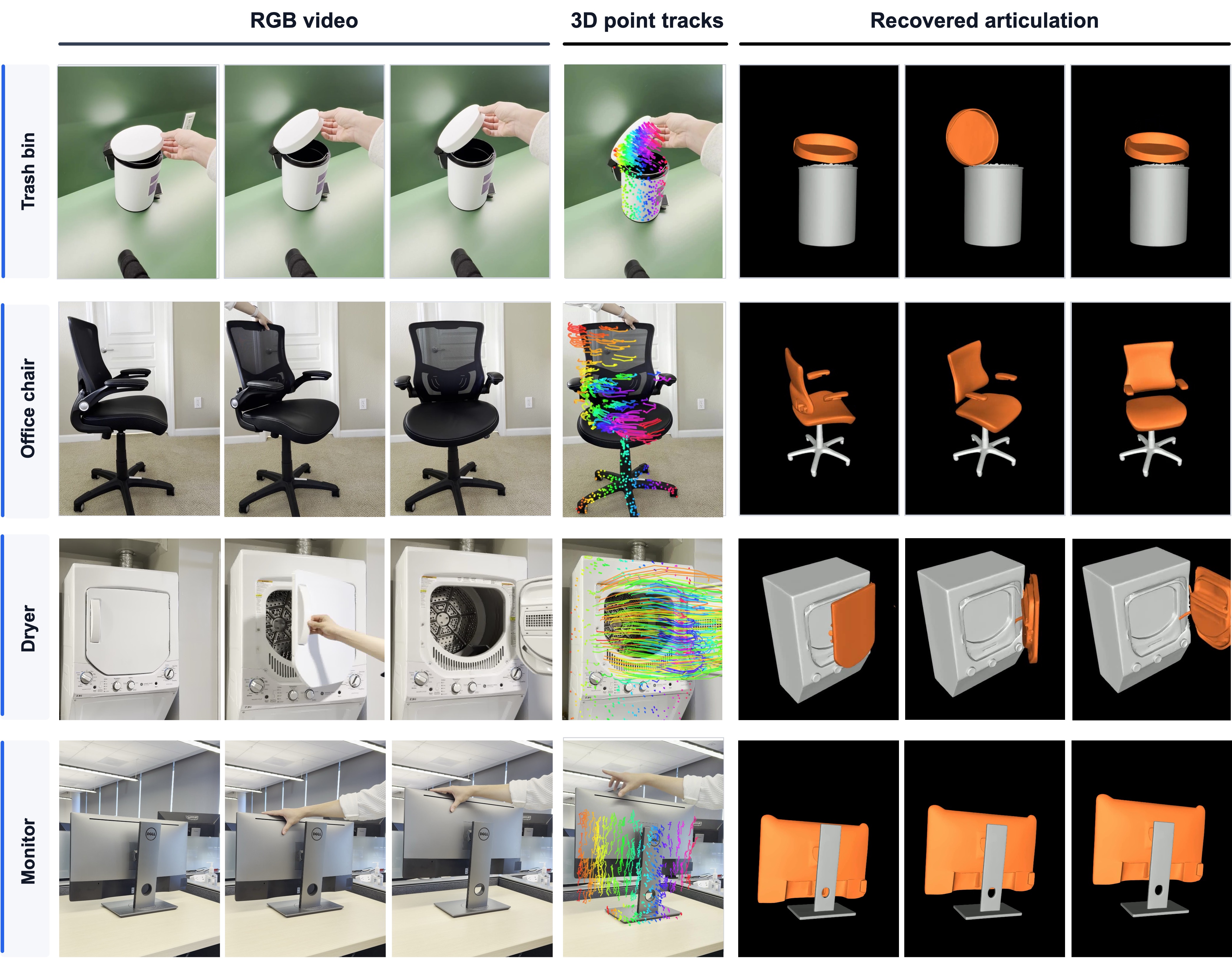}
    \caption{\textbf{Articulation recovery from real human videos.} From left to right, we show the RGB video input, 3D point tracks overlaid on an intermediate frame, and the reconstructed articulated asset. Our method handles both revolute and prismatic objects without any sensor depth or a prior object model.}
    \label{fig:qualvis}
    \vspace*{-0.2cm}
\end{figure*}
\subsection{Hand Alignment and Re-targeting}
\label{sec:retargeting}

HaWoR~\cite{zhang2025hawor} reconstructs a hand-mesh trajectory
$\mathcal{H}_{1:T}=\{\mathbf{x}^{h}_{j,t}\}$ in its estimated world frame,
while the articulation pipeline represents the moving-link trajectory in the
canonical object frame as
$\mathbf{T}^{m}_{t}=g_{\hat y}(q_t;\hat{\boldsymbol{\theta}})$. Although both are recovered from the same video, their independently estimated scale, pose, and depth can leave the hand displaced from the object. The main challenge is therefore to align the two trajectories while preserving their relative motion.

Since the hand is not visible across all frames and HaWoR sometimes confuses distance changes with scale changes, we first perform scale normalization of hand anatomy to the average value, then apply linear interpolation to make up for the frames where HaWoR fails to detect a hand. Thirdly, we align the depth of the hand such that it has the most intersection or the least distance to the generated mesh. The scale is simultaneously adjusted by preserving the same back-projected silhouette.

We picked contact-relevant hand vertices and track
$\mathbf{z}^{*}_{1:T}$ with position actuators in MuJoCo. The recovered object joint remains passive: its trajectory $q_{1:T}$ is used to align the
demonstration and define the target state, but it is never commanded during a
rollout. Any object motion must therefore arise from simulated hand contact.

Because our experiments use a human-shaped simulated hand, this kinematic
mapping is sufficient for tasks that do not require grasping. The recovered hand trajectory,
object trajectory, contact locations, and articulated asset also provide the
inputs needed by cross-embodiment retargeting methods. For example, re-targeting approaches like SPIDER~\cite{pan2025spider} and DexMachina~\cite{zhao2025dexmachina} refine kinematic references into dynamically feasible robot trajectories using
physics-based sampling and contact guidance. Such methods can replace our kinematic mapping when re-targeting the reconstructed interaction to a robot rather than the simulated human hand. We discuss additional details on physics-based re-targeting with articulated objects in the website.


\begin{table*}[t]
    \centering
    \caption{\textbf{Articulated joint estimation on simulation videos.} This evaluation uses simulator-derived depth, tracks, and masks, and analyzes the articulation optimization instead of the full pipeline. (n/a) means the method fails to produce any result at given condition.}
    \label{tab:simulation_articulation}
    \setlength{\tabcolsep}{10pt}
    \renewcommand{\arraystretch}{1.15}
    \begin{tabular}{clccc}
        \toprule
        & Method & Type acc. $\uparrow$ & Axis/dir. err. ($^\circ$) $\downarrow$ & Axis loc. err. (mm) $\downarrow$ \\
        \midrule
        \multirow{2}{*}{\rotatebox[origin=c]{90}{\small w/ Orbit}}
          & VideoArtGS~\cite{liu2025videoartgs}   & 0.93 & $1.81 \pm 0.74$ & $2.44 \pm 3.86$ \\
          & Articulat3D~\cite{guo2026articulat3d} & 0.91 & $1.53 \pm 1.17$ & $2.58 \pm 0.99$ \\
        \cmidrule(lr){2-5}
        \multirow{4}{*}{\rotatebox[origin=c]{90}{\small w/o Orbit}}
          & VideoArtGS & n/a  & n/a               & n/a \\
          & Articulat3D & 0.85 & $6.15 \pm 9.03$   & $7.82 \pm 8.16$ \\
          & Articulate-Anything~\cite{le2024articulateanything} & 0.66 & $30.66 \pm 25.92$ & $79.30 \pm 125.58$ \\
          & Ours                                                & \textbf{0.97} & $\mathbf{1.96 \pm 0.45}$ & $3.93 \pm 1.01$ \\
        \bottomrule
    \end{tabular}
\end{table*}
\newpage
\section{Experiments}
\label{sec:experiments}

    Through experiments, we aim to understand two questions: 1) Can our approach recover the articulation observed in an RGB video, and 2) can the aligned 3D hand trajectory use the recovered asset to reproduce the interaction in physics simulation?

\subsection{Experimental Setup}
\label{sec:exp_setup}

\paragraph{Real videos}
We evaluate the full pipeline on eight casual RGB videos spanning laptop, hinged container, dryer, stapler, monitor, puncher, and office chair. Each contains one dominant revolute or prismatic interaction and is processed independently. Our approach receives only the RGB frames from the respective videos as input, and no additional auxiliary information.

\paragraph{Simulation videos}
For quantitative articulation evaluation, a separate set of simulated models under known revolute and prismatic motion is chosen from the PARIS dataset~\cite{liu2023paris}. To exclude the performance coupling across different modules, the depth map, point tracks, and segmentation masks are directly derived from the ground truth mesh. These simulator annotations provide ground truth joint parameters that are otherwise not available from casual real videos, and thus allow us to compare against the respective ground truths.  We selected 10 simulated objects in PARIS, and for each joint, we drive the joint with 7 different combinations of motion range and starting pose to mimic the variability in real articulated objects. 

\paragraph{Baselines} For articulation recovery, we compare with Articulate-Anything~\cite{le2024articulateanything}, VideoArtGS~\cite{liu2025videoartgs}, and Articulat3D~\cite{guo2026articulat3d}, which cover a VLM-based asset pipeline and the two closest monocular-video digital-twin methods. We do not include methods requiring an RGB-D capture or 4D point clouds in the main RGB-only table~\cite{kerr2024rsrd,peng2025video2articulation,werby2025artipoint,liu2023reart}.

\subsection{Evaluation of the Full Framework from Human Videos to Articulated Objects}
\label{sec:qual_results}

Figure~\ref{fig:qualvis} shows the full results of our framework for mesh estimation and recovering the articulation from real RGB human videos. We can see that the inferred mesh parts and articulation are both plausible and accurately reflect the intended motion of the object from the respective videos. We also evaluate the baselines on the real videos and find that our modular approach is significantly better in estimating the joints and also recovering the full interactable articulated object. We provide these additional qualitative comparisons in the website.

\subsection{Evaluation of Articulation Only}
\label{sec:eval_articulation}

We isolate articulation recovery on the simulated objects, where every link and joint is known. We report joint-type accuracy, the sign-invariant angular error of the predicted axis or translation direction, and revolute-axis location error.

\begin{figure}[t]
    \centering
    \includegraphics[width=0.9\linewidth]{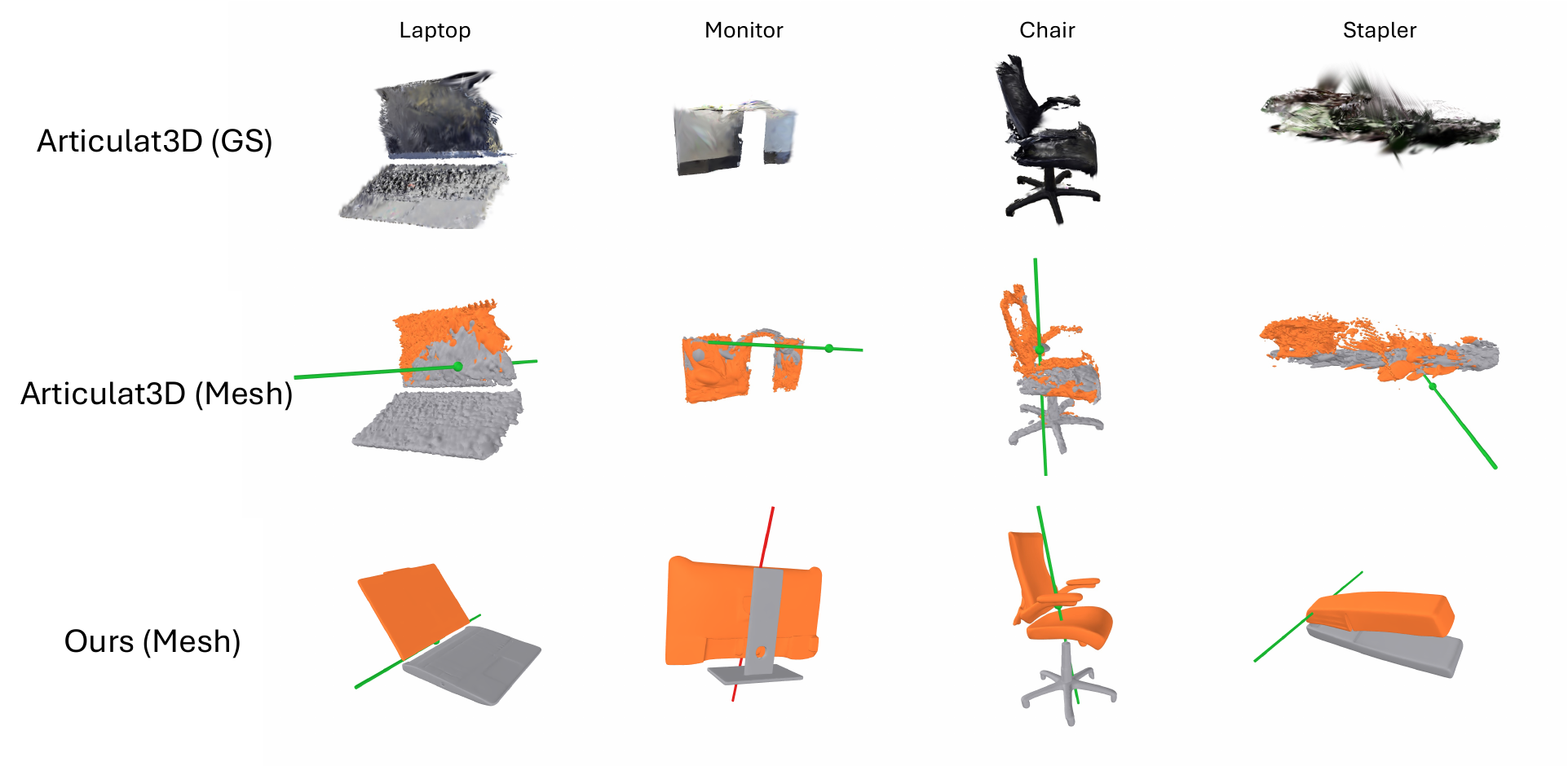}
    \caption{\textbf{Qualitative results for articulated joint estimation on real data for our approach compared with Articulat3D.} Each column shows the inferred joint for different objects. }
    \label{fig:qualvissim}
\end{figure}

This evaluation uses simulator-derived depth, tracks, and masks, and analyzes the articulation optimization instead of the full pipeline. Within this setting, the key distinction is camera motion: VideoArtGS operates only on images captured while the camera orbits the object, and Articulat3D degrades substantially without an orbiting view (Table~\ref{tab:simulation_articulation}). Since removing this constraint is one of the core contributions of our pipeline, we evaluate both baselines under two settings: 1) with orbit views, to report their best-case performance, and 2) without, for a head-to-head comparison against our method. With casual RGB sequences, our method significantly outperforms both baselines; on real human videos that violate the orbit assumption, they often fail to recover a usable articulated object altogether, as shown qualitatively in Figure~\ref{fig:qualvissim}. Against the orbit-view upper bound, our joint parameters are slightly worse, but the absolute axis orientation and translation errors remain small relative to a mesh spanning tens of centimeters. 

Figure~\ref{fig:qualvissim} also shows two examples where our method could successfully estimate the articulation and mesh, but Articulat3D fails because of lacking orbit view image as inputs.

\subsection{Hand--Object Interaction in Simulation}
\label{sec:eval_simulation}

The preceding evaluations establish the quality of the reconstructed object geometry and recovered articulation. We now aim to understand how well the reconstructed hand trajectory can be aligned with this asset and feasibly re-targeted to reproduce the demonstrated interaction in a simulator. We consider opening, closing, pushing, pulling, and folding interactions that can be completed without requiring a stable grasp. The reconstructed human hand is aligned with the recovered object and kinematically re-targeted to a simulated human hand in MuJoCo. The hand trajectory is replayed open loop, while the object joint remains passive, so all object motion in the simulator must arise from contact with the hand.

\begin{figure}[t]
    \centering
    \includegraphics[width=0.95\linewidth]{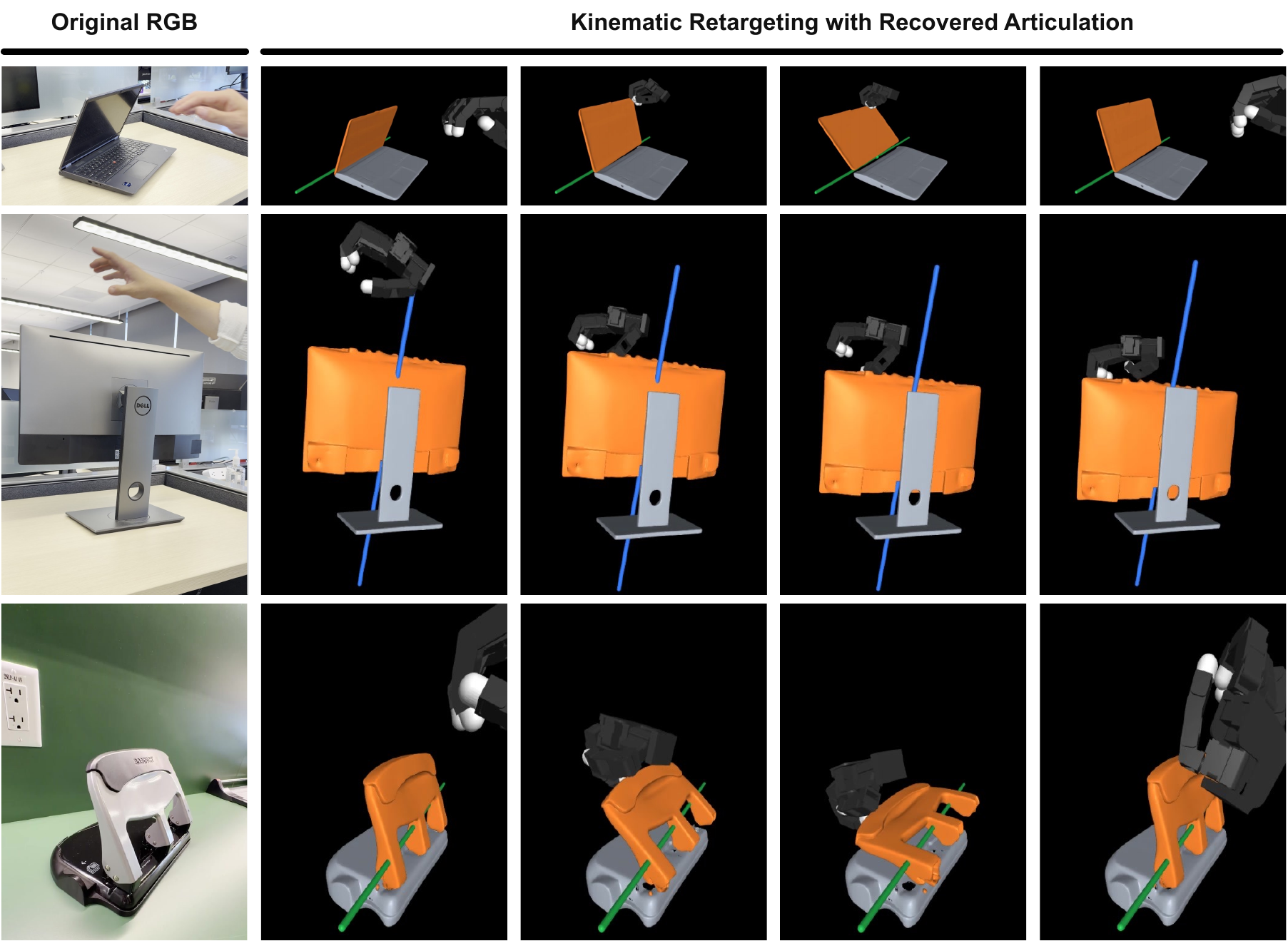}
\caption{\textbf{Interactive hand re-targeting in MuJoCo.}
We align the reconstructed human hand with the articulated asset and re-target its motion to a simulated robot hand. Results show that the re-targeted hand, when actuated, is able to plausibly manipulate the object.}
    \label{fig:interaction_sim}
\end{figure}

\begin{wraptable}{r}{0.47\textwidth}
    \centering
    \normalfont\normalsize
    \setlength{\tabcolsep}{4pt}
    \renewcommand{\arraystretch}{1.05}

    \captionsetup{
        font=normalsize,
        justification=raggedright,
        singlelinecheck=false,
        skip=6pt
    }

    \caption{\footnotesize
        \textbf{Contact-driven interaction after re-targeting.}
        We test whether the aligned and re-targeted hand can reproduce the demonstrated joint-state change in simulation, which is produced through contact and normalized against the demonstrated joint range.
    }
    \label{tab:interaction_sim}

    \begin{tabular}{@{}lcc@{}}
        \toprule
        Interaction
            & \shortstack{Joint progress\\(\%) $\uparrow$}
            & \shortstack{Final $q$ err.\\(\%) $\downarrow$} \\
        \midrule
        Laptop  & 63 & 35 \\
        Monitor & 39 & 59 \\
        Stapler & 92 & 4  \\
        Puncher & 97 & 1  \\
        \midrule
        Mean    & 73 & 25 \\
        \bottomrule
    \end{tabular}
\end{wraptable}
Figure~\ref{fig:interaction_sim} shows the respective real video and the corresponding initial, intermediate, and final MuJoCo states. This evaluation directly probes the quality of the hand--object alignment: an inaccurate relative pose, scale, or trajectory would cause the simulated hand to miss the object, or attempt to move it in the wrong direction. We report normalized joint progress, final joint-state error relative to the demonstrated range. Table~\ref{tab:interaction_sim} shows that the aligned hand trajectory can reliably interact with the recovered object in simulation.  These results establish the feasibility of aligning and re-targeting the recovered hand motion while preserving the contact needed to articulate the reconstructed object.


\newpage
\section{Discussion}
\label{sec:discussion_conclusion}

We presented a \textit{modular} real-to-sim framework that recovers an articulated object and aligned 3D hand trajectory from a casual monocular human video. By connecting pretrained predictions through multi-frame registration, motion-grounded joint fitting, and reprojection verification, the method produces an interactive MuJoCo asset without requiring sensor depth, prior object geometry, joint annotations, or robot demonstrations.

This work explores how far current best pre-trained vision models can take us toward recovering interactive articulated objects from human videos without the use of vision-language models. Our results show that combining visual predictions with geometric constraints and optimization can support both accurate articulation recovery and contact-driven hand replay in the settings studied here. These findings highlight the capabilities of existing vision models when their predictions are connected through the geometry and motion of the observed interaction.

The method remains limited by errors in monocular depth, tracking, segmentation, and single-image reconstruction, particularly under heavy occlusion or small joint motion. It currently handles one dominant revolute or prismatic joint and interactions in simulation that do not involve complex grasps; compound articulation, and deformable objects require richer contact and kinematic models. Looking ahead, extending our modular framework to recover more complex articulation and richer contact information from videos, together with advances in physics-based simulators, can enable dexterous manipulation across a wider range of objects.


{
    \small
    \bibliographystyle{ieeenat_fullname}
    \bibliography{main}
}

\clearpage
\setcounter{page}{1}
\maketitlesupplementary

\section{Simulated Dataset}
\label{sec:qualitative}
We use the PARIS dataset~\cite{liu2023paris} to construct synthetic sequences for comparing joint estimation performance among VideoArtGS~\cite{liu2025videoartgs}, Articulat3D~\cite{guo2026articulat3d}, and our method. PARIS provides ground-truth joint parameters and meshes at the initial and final articulation states, with moving and static parts segmented. We interpolate between these states to generate complete articulation sequences, then render RGB images, depth maps, and segmentation masks from specified camera poses. We also obtain ground-truth point tracks by recording the trajectories of a fixed set of mesh vertices throughout each sequence.

\begin{figure}[htbp]
    \centering
    \includegraphics[width=\linewidth]{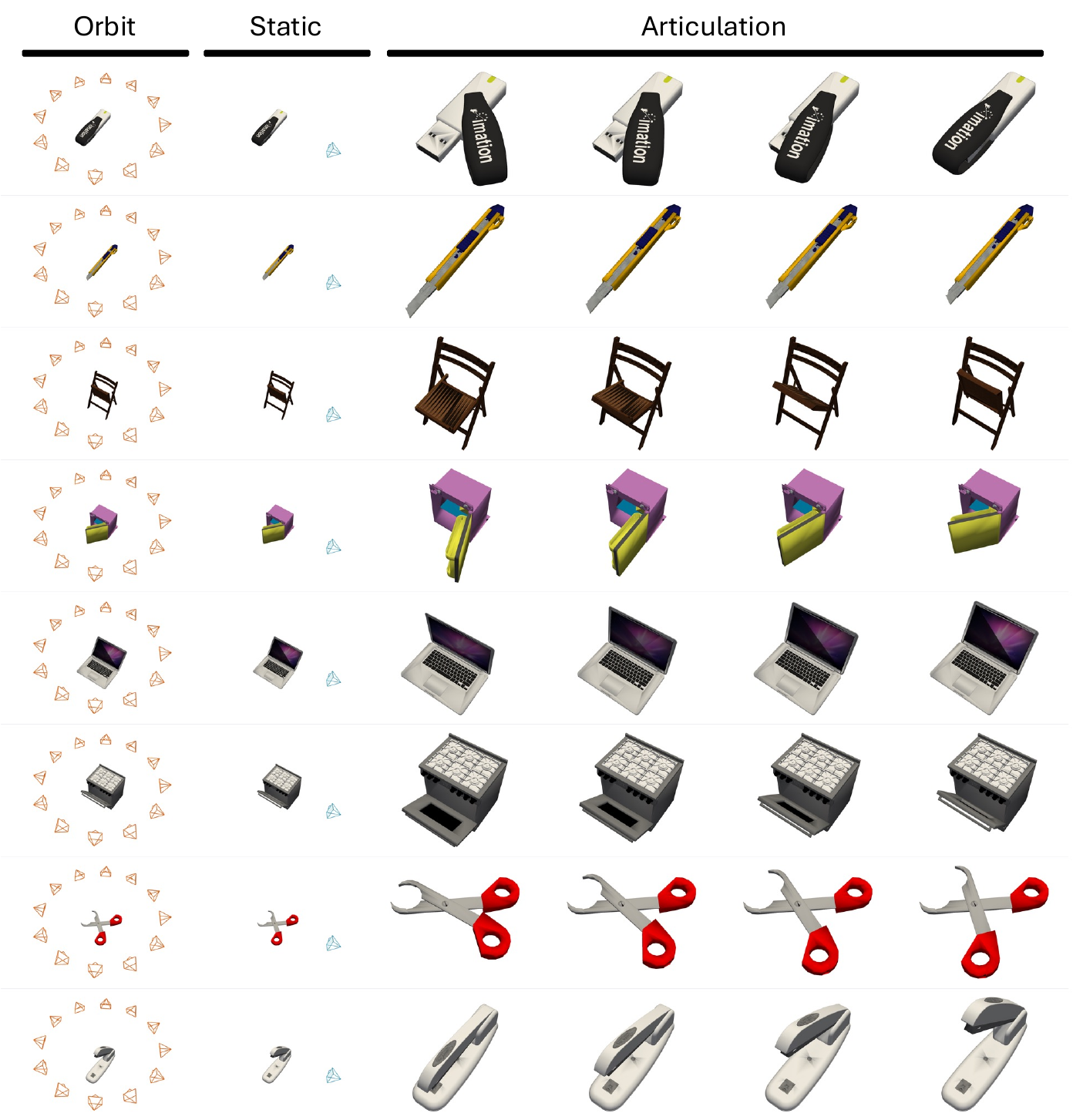}
    \caption{\textbf{Synthetic dataset for benchmarking the joint estimation method.} There are two different camera configurations: orbit camera and static camera. The actual articulation sequences are interpolated between the given initial and final states of the moving part.}
    \label{fig:sim}
\end{figure}

As shown in Figure~\ref{fig:sim}, we consider two camera configurations: an orbiting camera and a static camera. We use the orbiting configuration to evaluate VideoArtGS and Articulat3D under their intended camera-motion conditions. Because our method does not require camera motion, we report its joint estimation accuracy under the static-camera configuration.

\section{Joint Estimation}
\paragraph{Track-based joint initialization}
Given a video, per-frame part masks, camera parameters, segmented reference meshes, and 3D point trajectories of the moving part, we estimate its joint type, joint parameters, and articulation states. We select keyframes spanning diverse mask configurations within the available tracking window and reserve a subset for joint-type selection.

Let $\mathbf{x}_{i,t}$ denote the position of tracked point $i$ at frame $t$, and let $t_0$ be the mesh reference frame. We use $\mathbf{x}_{i,t_0}$ as the reference point positions, accounting for any difference between the tracker and mesh reference frames. For each training keyframe, we estimate a rigid transformation
by solving
\begin{equation}
    (R_t,\mathbf{d}_t)
    =
    \arg\min_{R\in SO(3),\,\mathbf{d}}
    \sum_{i\in\mathcal{I}_t}
    \left\|
        R\mathbf{x}_{i,t_0}+\mathbf{d}-\mathbf{x}_{i,t}
    \right\|_2^2.
\end{equation}
We solve this alignment using Kabsch registration with iterative residual trimming to reduce the influence of unreliable tracks. Non-finite correspondences are excluded, and registration residuals provide confidence weights for subsequent joint fitting.

\paragraph{Joint parameterization and fitting}
We fit both prismatic and revolute hypotheses to the estimated rigid transformations. Their motion models are
\begin{align}
    \mathcal{T}_{\mathrm{P}}(\mathbf{x};q_t)
    &= \mathbf{x}+q_t\mathbf{a}, \\
    \mathcal{T}_{\mathrm{R}}(\mathbf{x};q_t)
    &= \mathbf{c}
       + \exp\!\left(q_t[\mathbf{a}]_\times\right)
         (\mathbf{x}-\mathbf{c}),
\end{align}
where $\mathbf{a}$ is a unit axis, $\mathbf{c}$ is a point on the revolute axis, and $q_t$ denotes translation distance or rotation angle. For the prismatic hypothesis, we initialize $\mathbf{a}$ from the dominant direction of the estimated translations. For the revolute hypothesis, we initialize the axis from relative rotation vectors and recover an axis point through
\begin{equation}
    (I-R_t)\mathbf{c}\approx\mathbf{d}_t.
\end{equation}
We remove the axis-point ambiguity by constraining $\mathbf{c}$ to the plane through the reference point-cloud centroid perpendicular to $\mathbf{a}$. Random-subset hypothesis generation and robust residual scoring identify an initial consensus set. We then refine each hypothesis using confidence-weighted Huber least squares on rotation and translation residuals, normalizing translation by the reference point cloud's bounding-box diagonal.

\paragraph{Image-space refinement}
Starting from the provided object pose, we align the reference meshes by optimizing a shared rigid transformation and anisotropic scale against the static-part masks. We extract DINOv2 features, compress them using a shared PCA projection, and attach reference-frame descriptors to visible mesh vertices. These descriptors remain fixed on the mesh during articulation.

For each initialized joint hypothesis, we render the articulated mesh and compare its silhouette and features with the observations:
\begin{equation}
    \mathcal{L}_{\mathrm{img}}^t
    =
    \lambda_m\mathcal{L}_{\mathrm{mask}}^t
    + \lambda_b\mathcal{L}_{\mathrm{boundary}}^t
    + \lambda_f\mathcal{L}_{\mathrm{feature}}^t.
\end{equation}
The mask term penalizes missing foreground and rendered foreground outside the observed mask, with a reduced penalty on the latter. The boundary term measures distances from rendered silhouette edges to the observed mask boundary. The feature term measures cosine distance between rendered and observed descriptors over their valid foreground overlap.

We optimize the articulation states first, then jointly refine the states and joint parameters. A final stage additionally optimizes a shared rigid transformation and isotropic scale correction of both meshes:
\begin{equation}
    \mathcal{L}
    =
    \frac{1}{|\mathcal{K}_{\mathrm{tr}}|}
    \sum_{t\in\mathcal{K}_{\mathrm{tr}}}
    \left(
        \mathcal{L}_{\mathrm{img,mov}}^t
        + \tfrac{1}{2}\mathcal{L}_{\mathrm{img,stat}}^t
    \right)
    + \lambda_s\mathcal{L}_{\mathrm{smooth}}
    + \lambda_{\mathrm{scale}}(\log s)^2,
\end{equation}
where $s$ is the scale correction. The smoothness term penalizes second differences of the temporally ordered keyframe states:
\begin{equation}
    \mathcal{L}_{\mathrm{smooth}}
    =
    \frac{1}{K-2}
    \sum_{i=2}^{K-1}
    \left(q_{i+1}-2q_i+q_{i-1}\right)^2.
\end{equation}
The reference state is fixed to zero. The 3D trajectories are used only for initialization and do not contribute residuals to this image-space refinement.

\end{document}